%% file: icml_main.tex
\documentclass{article}

\usepackage{microtype}
\usepackage{graphicx}
\usepackage{amsmath}
\usepackage{booktabs} 
\usepackage[hyphens]{url}
\usepackage{ragged2e}
\usepackage{adjustbox}
\usepackage{wasysym}
\usepackage{subcaption}
\newcommand{\wcircle}{{$\RIGHTcircle$}}
\newcommand{\bcircle}{{$\CIRCLE$}}
\newcommand{\ecircle}{{$\Circle$}}

\usepackage{etoolbox}

\newcommand*\rot{\rotatebox{25}}

\usepackage{hyperref}

\usepackage[accepted]{icml2024-arxiv}

\usepackage{amssymb}
\usepackage{mathtools}
\usepackage{amsthm}

\usepackage{blindtext}

\usepackage[capitalize,noabbrev]{cleveref}

\theoremstyle{plain}
\newtheorem{theorem}{Theorem}[section]
\newtheorem{proposition}[theorem]{Proposition}
\newtheorem{lemma}[theorem]{Lemma}
\newtheorem{corollary}[theorem]{Corollary}
\theoremstyle{definition}
\newtheorem{definition}[theorem]{Definition}
\newtheorem{assumption}[theorem]{Assumption}
\theoremstyle{remark}

\usepackage{booktabs} 
\usepackage{tabularx}
\usepackage{adjustbox}
\usepackage{framed}
\usepackage{enumitem}
\usepackage{ragged2e}
\newcolumntype{P}[1]{>{\RaggedRight\hspace{0pt}}m{#1}}
\usepackage{color, colortbl}
\definecolor{Gray}{gray}{0.94}
\usepackage{pifont}
\usepackage[capitalize]{cleveref}
\usepackage{tcolorbox}
\usepackage{xspace}
\usepackage{xcolor}
\tcbuselibrary{skins}
\usepackage{caption}
\usepackage{mwe}
\usepackage{caption}
\usepackage{subcaption} 
\usepackage{tikz}
\usepackage{float}

\def\Snospace~{\S{}}

\usepackage[textsize=tiny]{todonotes}

\input{macros}

\input{std-macros}

\icmltitlerunning{\thepage $\quad\cdot\quad$A Safe Harbor for AI Evaluation and Red Teaming}

\begin{document}

\twocolumn[
\icmltitle{A Safe Harbor for AI Evaluation and Red Teaming}



\icmlsetsymbol{equal}{*}
\icmlsetsymbol{equaly}{**}

\begin{icmlauthorlist}
\icmlauthor{Shayne Longpre\textsuperscript{*}}{mit}
\icmlauthor{Sayash Kapoor\textsuperscript{**}}{princeton}
\icmlauthor{Kevin Klyman\textsuperscript{**}}{stanford}
\icmlauthor{Ashwin Ramaswami}{georgetown}
\icmlauthor{Rishi Bommasani}{stanford}
\icmlauthor{Borhane Blili-Hamelin}{airva}
\icmlauthor{Yangsibo Huang}{princeton}
\icmlauthor{Aviya Skowron}{eai}
\icmlauthor{Zheng-Xin Yong}{brown}
\icmlauthor{Suhas Kotha}{cmu}
\icmlauthor{Yi Zeng}{vt}
\icmlauthor{Weiyan Shi}{northeastern}
\icmlauthor{Xianjun Yang}{ucsb}
\icmlauthor{Reid Southen}{}
\icmlauthor{Alexander Robey}{up}
\icmlauthor{Patrick Chao}{up}
\icmlauthor{Diyi Yang}{stanford}
\icmlauthor{Ruoxi Jia}{vt}
\icmlauthor{Daniel Kang}{uiuc}
\icmlauthor{Sandy Pentland}{mit}
\icmlauthor{Arvind Narayanan}{princeton}
\icmlauthor{Percy Liang}{stanford}
\icmlauthor{Peter Henderson}{princeton}
\\\vspace{0.3cm}
\icmlauthor{\textmd{March 5, 2024}}{}
\end{icmlauthorlist}

\icmlaffiliation{mit}{MIT}
\icmlaffiliation{princeton}{Princeton University}
\icmlaffiliation{stanford}{Stanford University}
\icmlaffiliation{airva}{AI Risk and Vulnerability Alliance}
\icmlaffiliation{georgetown}{Georgetown University}
\icmlaffiliation{eai}{Eleuther AI}
\icmlaffiliation{brown}{Brown University}
\icmlaffiliation{cmu}{Carnegie Mellon University}
\icmlaffiliation{vt}{Virginia Tech}
\icmlaffiliation{northeastern}{Northeastern University}
\icmlaffiliation{ucsb}{UCSB}
\icmlaffiliation{up}{University of Pennsylvania}
\icmlaffiliation{uiuc}{UIUC.  \textsuperscript{*} Corresponding author.  \textsuperscript{**} Equal contribution}

\icmlcorrespondingauthor{Shayne Longpre}{slongpre@media.mit.edu}
\icmlauthorrunning{}

\icmlkeywords{Machine Learning, ICML}

\vskip 0.3in
]



\printAffiliationsAndNotice{}  

\input{icml_sections/00_abstract}
\input{icml_sections/01_introduction}
\input{icml_sections/02_evaluation}
\input{icml_sections/03_challenges}
\input{icml_sections/04_safe_harbors}

\input{icml_sections/05_related_proposals}
\input{icml_sections/06_conclusion}
\section*{Acknowledgements}

We would like to thank Stephen Casper, Dylan Hadfield-Menell, Yacine Jernite, Amit Elazari, and Harley Geiger for their insightful feedback and guidance.

\bibliography{references}
\bibliographystyle{icml2024}

\newpage
\appendix
\renewcommand{\thefigure}{A\arabic{figure}}
\renewcommand{\thetable}{A\arabic{table}}

\setcounter{figure}{0}
\setcounter{table}{0}
\onecolumn
\section*{Appendix}

\input{icml_sections/10_appendix}

\end{document}

%% file: macros.tex
\newcommand\tldr[1]{\textcolor{gray}{\textit{[#1]} \\}}
\newcommand\tldrDone[1]{}
\newcommand\todocite[1]{\textcolor{green}{[Add cite: #1]}}

\newcommand\kk[1]{\textcolor{olive}{[KK: #1]}}
\newcommand\srl[1]{\textcolor{violet}{[SL: #1]}}
\newcommand\an[1]{\textcolor{blue}{[AN: #1]}}

\renewcommand\tldr[1]{}
\renewcommand\tldrDone[1]{}
\renewcommand\todocite[1]{}
\renewcommand\kk[1]{}
\renewcommand\srl[1]{}
\renewcommand\an[1]{}


%% file: std-macros.tex
\ifthenelse{\isundefined{\definition}}{\newtheorem{definition}{Definition}}{}
\ifthenelse{\isundefined{\assumption}}{\newtheorem{assumption}{Assumption}}{}
\ifthenelse{\isundefined{\hypothesis}}{\newtheorem{hypothesis}{Hypothesis}}{}
\ifthenelse{\isundefined{\proposition}}{\newtheorem{proposition}{Proposition}}{}
\ifthenelse{\isundefined{\theorem}}{\newtheorem{theorem}{Theorem}}{}
\ifthenelse{\isundefined{\lemma}}{\newtheorem{lemma}{Lemma}}{}
\ifthenelse{\isundefined{\corollary}}{\newtheorem{corollary}{Corollary}}{}
\ifthenelse{\isundefined{\alg}}{\newtheorem{alg}{Algorithm}}{}
\ifthenelse{\isundefined{\example}}{\newtheorem{example}{Example}}{}

%% file: icml_sections/00_abstract.tex
\begin{abstract}
Independent evaluation and red teaming are critical for identifying the risks posed by generative AI systems.
However, the terms of service and enforcement strategies used by prominent AI companies to deter model misuse have disincentives on good faith safety evaluations.
This causes some researchers to fear that conducting such research or releasing their findings will result in account suspensions or legal reprisal.
Although some companies offer researcher access programs, they are an inadequate substitute for independent research access, as they have limited community representation, receive inadequate funding, and lack independence from corporate incentives.
We propose that major AI developers commit to providing a legal and technical safe harbor, indemnifying public interest safety research and protecting it from the threat of account suspensions or legal reprisal.
These proposals emerged from our collective experience conducting safety, privacy, and trustworthiness research on generative AI systems, where norms and incentives could be better
aligned with public interests, without exacerbating model misuse.
We believe these commitments are a necessary step towards more inclusive and unimpeded community efforts to tackle the risks of generative AI.
\end{abstract}

%% file: icml_sections/01_introduction.tex
\section{Introduction}
\label{sec:introduction}

Generative AI systems have been deployed rapidly in recent years, amassing hundreds of millions of users.
These systems have already raised concerns for widespread misuse, bias \citep{deshpande2023toxicity}, hate speech \citep{chatbot_hatespeech}, privacy concerns \citep{carlini2021extracting, carlini2023extracting}, disinformation \citep{burtell2023artificial}, self harm \citep{park2023supporting}, copyright infringement \citep{henderson2023foundation, copyright_1}, fraud \citep{stupp2019fraudsters}, weapons acquisition \citep{boiko2023emergent,urbina2022dual}, and the proliferation of non-consensual and abusive images \citep{lakatos2023revealing,thiel_generative_2023}, among others \citep{kapoor2024societal}.
To ensure sufficient public scrutiny and accountability, such high-impact systems should be evaluated \citep{liang2023holistic, solaiman2023evaluating, weidinger2023sociotechnical} by \emph{independent and external} entities \citep{raji2022audit, birhane2024ai}.
Despite this, leading generative AI companies provide limited transparency and access into their systems, with transparent audits showing only 25\% of policy enforcement and evaluation criteria were satisfied on average \citep{bommasani2023foundation}; and with no company providing reproducible evaluations to characterize the effectiveness of their risk mitigations.

\input{tables/terminology}

Leading AI companies' terms of service prohibit independent evaluation into most sensitive model flaws (see \autoref{tab:policies}).
While these terms act as a deterrent to malicious behavior, they also restrict good faith research---auditors fear that releasing findings or conducting research could lead to their accounts being suspended, ending their ability to do such research, or even lawsuits for violating the terms of service.
Already, in the course of conducting good faith research, researchers' accounts have been suspended without warning, justification, or an opportunity to appeal~\citep{marcus2024generative}.
While some companies authorize selected research through researcher access programs, their community representation remains limited and lacks independence from corporate incentives such as favoritism towards researchers aligned with the company's values.
Together, these observations stoke concerns that generative AI companies could emulate the transparency and accountability challenges with social media platforms---limiting researcher transparency and access can mitigate dangerous headlines, public relations fallout, and lawsuits, but at the expense of public interests \citep{abdo2022safe,diresta2022blackbox}.

As a group of researchers whose expertise spans AI red teaming, safety, and evaluation, as well as privacy, security, and the law, we have experienced first hand the negative effects of legal uncertainty and technical barriers to conducting important research (\cref{tab:interviews}).
To improve the status quo, we propose that generative AI companies commit to two protections for independent public interest research. 
First, AI companies should provide a \textbf{legal safe harbor} by offering legal protections for good faith research, provided it is conducted in line with vulnerability disclosure policies (as defined in \cref{tab:terminology}). 
Second, companies should provide a \textbf{technical safe harbor}, protecting safety researchers from having their accounts subject to moderation or suspension. 
These are fundamental access requirements for inclusive evaluation of generative AI systems.
Building on prior work for algorithmic bug bounties \citep{Elazari2018,kenway2022bugs,raji2022audit} and  social media data access \citep{abdo2022safe}, we recommend ways to implement these protections for independent AI evaluation without undermining the processes that prevent model misuse.
Specifically we propose that companies delegate account authorization to trusted universities or nonprofits, or provide transparent recourse for accounts suspended in the course of research. 
These voluntary commitments align with the stated goals of AI companies: to support wider participation in AI safety research, minimize corporate favoritism, and encourage community safety evaluations (see \cref{sec:company-support}). 
We hope generative AI companies will adopt these commitments to establish better community norms, improve trust in their services, and bolster much needed AI safety in proprietary systems.

%% file: tables/terminology.tex
\begin{table*}[t!]
\centering
\footnotesize
\renewcommand{\arraystretch}{1.5}
\begin{adjustbox}{width=\textwidth}
\begin{tabular}{p{3.1cm} | p{14cm}}
\toprule
\textsc{Terminology} & \textsc{Context} \\
\midrule
\textbf{Usage Policy} & A company’s usage policy dictates what uses of its AI systems are acceptable or unacceptable. Usage policies generally prohibit inputs that elicit a range of undesirable model outputs, beyond what is already illegal. For example, see \href{https://console.anthropic.com/legal/aup}{Anthropic's Acceptable Use Policy}. \\ \rowcolor[gray]{0.9}
\textbf{Terms of Service} & A company's terms of service imposes legal rules on users of their services. Violations of the usage policy are violations of the terms of service and can be enforced by terminating accounts or taking legal action. \\
\parbox[t]{3.2cm}{\textbf{Generative AI \\ Evaluation \& Red \\ Teaming}} & In security fields, a \href{https://csrc.nist.gov/glossary/term/red_team}{red team} refers to a group authorized to emulate an adversary’s attack against an organization’s security systems.
This term has been adopted by the AI community to instead describe \href{https://csrc.nist.gov/glossary/term/penetration_testing}{penetration testing} of a broader set of system flaws than traditional security \citep{hpc2023legal}.
In this context, we are referring to testing of released systems by third party ethical hackers, who may or may have explicit consent.
\\ \rowcolor[gray]{0.9}
\textbf{Safe Harbor} & A safe harbor is a measure to provide legal protection to hackers engaged in ``good faith'' research, abiding by pre-agreed rules of engagement, or vulnerability disclosure policy (e.g. \citet{hackerone2023safeharbor}). \\
\textbf{Good Faith Research} & ``Good faith \emph{security} research means accessing a computer solely for purposes of good-faith testing, investigation, and/or correction of a security flaw or vulnerability, where such activity is carried out in a manner designed to avoid any harm to individuals or the public, and where the information derived from the activity is used primarily to promote security or safety...'' \citep{doj2022cfaa}. We generalize this definition to research beyond security, including soliciting any unwanted behavior in the AI system normally disallowed by the company's usage policy, which we broadly refer to as ``safety research'' in this work. \\ \rowcolor[gray]{0.9}
\parbox[t]{3.2cm}{\textbf{Vulnerability Disclosure Policy}} & A vulnerability disclosure policy establishes rules of engagement for third party ethical hackers. This includes disclosure requirements for discovered vulnerabilities, but also other mandatory protocols \citep{bugcrowd2023vulnerability}. 
\\
\textbf{Chilling Effects} & Chilling effects describe the inhibition or discouragement of important research, in this case due to a lack of legal and technical protections, as well as uncertain norms around AI evaluation and red teaming. \\
\bottomrule
\end{tabular}
\end{adjustbox}
\caption{We \textbf{define and contextualize the technical terminology} used in this work, which is often used in other disciplines.
}
\label{tab:terminology}
\vspace{-2mm}
\end{table*}

%% file: icml_sections/02_evaluation.tex
\section{Background \& Motivations}
\label{sec:eval}

Widely used online platforms can have significant socio-economic impact \citep{zuboff2023age,horwitz2021facebook}.
In this section we highlight three reasons to motivate new protections for independent research into generative AI platforms:
\begin{enumerate}
    \item Social media research has been burdened by a lack of transparency and access, with a rise in legal repercussions for journalism and academic research \citep{abdo2022safe,delong2021facebook, belanger2023researchers}.
    \item There is growing concern that widespread risks of generative AI will impact a wider swathe of society. Fostering wider participation in AI evaluation will require commitments to remove disincentives, obstacles, and favoritism in researcher access. 
    \item \emph{Independent} AI evaluation is increasingly vital to fair assessments of AI risks, and informed policy debates.
\end{enumerate} 

We expand on each of these below, using terminology we define in \cref{tab:terminology}.

\subsection{Avoiding the Fate of Social Media Platforms}

\paragraph{Prominent social media platforms block researcher access to the detriment of public interests.}
Civil societies and researchers argue that social media companies have systematically limited researcher access to their platforms, restricting journalism and creating a chilling effect on critical public interest research \citep{diresta2022blackbox, mozilla2023socialmedia, aspen2021commission, persily2021proposal}. Specifically, platforms wield their terms of service to gatekeep access to publicly posted data and limit negative public exposure from independent research. \citet{abdo2022safe} argue for ``a safe harbor for platform research,'' which would include legal provisions that protect researchers and journalists.
In the absence of such provisions, researchers have reported platform gatekeeping, account suspensions, cease-and-desist letters and general fears of liability in the course of public interest research, which have resulted in chilling effects \citep{delong2021facebook, barclay2021facebook, belanger2023researchers}.
The computer and internet security fields have also seen contentious legal threats and lawsuits against academics \citep{greene2001sdmi, brodkin2021missouri, disclose2021research}, resulting in new guidelines from the United States Department of Justice that ``good-faith security research should not be charged'' \citep{doj2022cfaa}.
Companies building generative AI models have the opportunity to protect good faith research before harm from their systems becomes as widespread as that from social media.

\paragraph{Conducting research on generative AI comes with additional challenges compared to social media.} 
Compared to past digital technologies, prominent models require accounts to be used (unlike search engines), and their outputs are not publicly visible (unlike posts on many social media platforms)~\cite{narayanan2023transparencyreports}. 
These factors provide developers with comparatively greater control over who accesses their systems, which could exacerbate gatekeeping.
The lack of transparency from top developers compounds this issue, with little information available about how and where generative AI systems are used, and to what end \citep{bommasani2023transparency}. 
For external researchers, the models themselves are also black boxes, as developers often do not disclose model architectures, sizes, or training data. 
This limits independent research to evaluate the risks, capabilities, safety, and societal impact of generative AI \citep{casper2024blackbox}.  

\subsection{The Importance of Independent AI Evaluation}
\label{sec:independent-eval}

\paragraph{Concerns over the risks and harms of generative AI are mounting.}
Today, AI systems like ChatGPT have amassed over 100 million weekly users \citep{hu2023100mn}, exceeding the growth rate of social media platforms.
Generative AI systems have already exhibited ``unsafe'' behavior---generating highly undesirable and even illegal content---attracting regulatory attention as a result. 
More specifically, generative AI systems can generate toxic content~\citep{deshpande2023toxicity}, libel, hate speech~\citep{chatbot_hatespeech}, and privacy leaks~\citep{carlini2021extracting, carlini2023extracting, li2023multi, huang2023privacy, nasr2023scalable}. They have also been used to scale disinformation \citep{burtell2023artificial}, fraud \citep{stupp2019fraudsters, ftc2023voice}, malicious tool usage \citep{li2023chatgpt, pa2023attacker, renaud2023chatgpt}, copyright infringement \citep{henderson2023foundation, copyright_1, copyright_2, shi2024detecting, longpre2023data}, non-consensual intimate imagery \citep{lakatos2023revealing}, and child sexual abuse material \citep{thiel_generative_2023}, as well as provide instructions for self-harm \citep{park2023supporting, xiang2023man}, acquiring weapons \citep{boiko2023emergent, bio_hazard}, and building weapons of mass destruction \citep{urbina2022dual, soice2023can}.
At the extreme end, even CEOs of AI model developers have speculated generative AI will upend labor markets \citep{suleyman2023coming} and even pose more severe risks \citep{barrabi2023sam, hendrycks2023overview}. These wide ranging concerns, from the developers themselves, motivate the need for protected independent access.

\input{tables/interviews}

\paragraph{Independent AI evaluation and red teaming are crucial for uncovering vulnerabilities, before they proliferate.}
Independent researchers often evaluate or ``red team'' AI systems for a broad range of risks. 
``Red teaming'', a subset of evaluation, has been adopted by the AI community as a term of art to describe these evaluations aimed at uncovering pernicious system flaws.
In this work, we refer specifically to red teaming of \emph{publicly released} AI systems (rather than pre-release testing), by \emph{external} researchers, rather than internal teams. 
Some companies do also provide internal or by-invitation pre-release red teaming, e.g. OpenAI.
While all types of testing are critical, external evaluation of AI systems that are already deployed is widely regarded as essential for ensuring safety, security, and accountability \citep{kenway2022bugs, anderljung2023frontier, raji2022audit}. 
Post-release, external red-teaming research has uncovered vulnerabilities related to low resource languages \citep{yong2023low}, conjugate prompting attacks \citep{kotha2023understanding}, adversarial prompts \citep{maus2023black,zou2023universal,robey2023smoothllm}, generation exploitation attacks \citep{huang2023catastrophic}, persuasion attacks \citep{xu2023earth, zeng2024johnny}, a wide range of jailbreaks \citep{wei2023jailbroken, shen2023anything, liu2023goal, zou2023universal, shah2023scalable}, text-to-image vulnerabilities \citep{parrish2023adversarial}, automatic red teaming \citep{ge2023mart, yu2023gptfuzzer,chao2023jailbreaking, zhao2024weaktostrong}, and undetectable methods for fine-tuning away safety mitigations within the platform APIs \citep{qi2023fine, yang2023shadow, zhan2023removing}.
See \cref{app:more-red-teaming} for additional examples.
These works illustrate how such research benefits AI companies: the research community assists in-house research teams by uncovering vulnerabilities, sharing findings and data, before systems cause major harm. 

\textbf{Independent AI evaluation provides impartial perspectives, that are necessary for informed regulation}
As the above examples have shown, independent research has uncovered unexpected flaws, aiding company efforts, and expanding the collective knowledge around both vulnerabilities and defenses.
These findings have informed the policy and regulatory discussions, including around the types of model vulnerabilities, and their comparative safety of open and closed foundation models \citep{narayanan2023model,lambert2023undoing}. 
However, as we shall see, it isn't clear that we are seeing the full benefits from a thriving red teaming ecosystem (\cref{sec:challenges}).

Without robust independent evaluation, companies' own developer safety teams may not be sufficiently large or diverse to fully represent the diversity of global users their products already serve, and the scale of risks they have acknowledged \citep{costanza2022auditauditors}.
While companies do invite third-party evaluators, there are well known conflicts of interest without independence in the auditor selection process \citep{moore2006conflicts}.
As the Ada Lovelace Institute and another dozen civil societies remarked at the recent AI Safety Summit in the UK, ``Companies cannot be allowed to assign and mark their own homework. Any research efforts designed to inform policy action around AI must be conducted with unambiguous independence from industry influence'' \citep{adalovelace2023postsummit}. 

%% file: tables/interviews.tex
\begin{table*}[t!]
\centering
\footnotesize
\renewcommand{\arraystretch}{1.5}
\begin{adjustbox}{width=\textwidth}
\begin{tabular}{p{3.1cm} | p{14cm}}
\toprule
\textsc{Theme} & \textsc{Observations} \\
\midrule
\parbox[t]{3.1cm}{\textbf{Chilling Effect on \\ Research}} & Safety research can result in companies suspending researcher access, citing terms of service violations. This can have broad chilling effects as access is critical for the other work conducted by these researchers.
Many researchers only begin their work after observing first-movers, and scope their practices to emulate those precedents. As a result, vital safety research may be delayed or circumscribed due to uncertainty and caution over account moderation outcomes. \\ \rowcolor[gray]{0.9}
\parbox[t]{3.1cm}{\textbf{Chilling Effect on \\ Vulnerability \\ Disclosure}} & It is unclear whether and how researchers should publicly release their findings, methods or the exploits themselves. In the absence of explicit guidance, they may be too broad or too limited in how they share their results, to the detriment of the community---for instance, by only sharing findings with a small group of other researchers, such as close personal contacts.
When researchers are overly cautious in sharing their work it frequently results in siloed research that is less reproducible, or delayed disclosure, especially around the most sensitive findings, which could be to the detriment of public awareness. \\
\parbox[t]{3.1cm}{\textbf{Incentives to Tackle the Wrong Problems}} & There is an incentive to prioritize less important risks as the focus of safety work both for the uncertainty of repercussions from the companies or community. For instance, researchers might choose to investigate more benign prompt attacks rather than more offensive or dangerous attacks, such as focusing on text rather than more evocative visual outputs, or tool usage.
\\ \rowcolor[gray]{0.9}
\parbox[t]{3.1cm}{\textbf{Favoritism and Imbalanced \\ Representation}} & Admission into researcher access programs and favorable responses to safety work can be dependent on connections to the companies. For instance, there is a strong impression that access to OpenAI employees improves access to their programs. External researchers who are not already well connected may not hear back at all from their applications or receive any justification for rejection, as no obligation currently exists on the part of AI companies. Part of this may be due to companies being backlogged with applications from researchers, having dedicated few resources to this task.
\citet{costanza2022auditauditors} point out the problems introduced by imbalanced auditor representation.
\\
\parbox[t]{3.1cm}{\textbf{Unclear \& \\ Undefined Norms}} & Impressions of basic norms and expectations vary widely, including with respect to appropriate threat models, whether and when to notify companies in advance of publication, what forms of red teaming are acceptable, and whether to release findings, methods, or prompts at all, or how to do so responsibly. Additionally, the type of API access, moderation policies, disclosure processes, and even the likelihood of response to a disclosure vary dramatically by company, leaving researchers without well-defined protocols that would enable them to confidently conduct important safety and security work. \\ \rowcolor[gray]{0.9}
\parbox[t]{3.1cm}{\textbf{A Choice Between Open and Closed Access}} & Researchers prefer to red team deployed systems that have millions of users and therefore pose immediate risks. However, effective and rigorous research requires deep access to the model \citep{casper2024blackbox}, which proprietary systems rarely provide. As \citet{friedler2023airedteaming} have noted, ``for red-teaming conducted by external groups to be effective, those groups must have full and transparent access to the system in question.''
In particular, researchers often require finer-grained access to internal model representations (e.g. ``logits''), access to both the base and aligned model, and continual access to a static model, without its API changing or becoming deprecated. Additionally, the underlying source of moderation in closed systems is difficult to diagnose: did the moderation endpoint catch an inappropriate user input, did the model itself abstain from answering, or did the user interface curtail an inappropriate response?
\\
\bottomrule
\end{tabular}
\end{adjustbox}
\caption{
\textbf{Themes and observations attributed to informal discussions among authors and colleagues working on AI evaluation and red teaming.} 
We describe the main challenges to conducting rigorous evaluations of widely used generative AI systems.}
\label{tab:interviews}
\vspace{-3mm}
\end{table*}

%% file: icml_sections/03_challenges.tex
\section{Challenges to Independent AI Evaluation}
\label{sec:challenges}

We first discuss the mixed incentives and uncertainty faced by red teaming researchers, followed by analysis of the existing researcher protections, access programs, and their limitations.

\textbf{AI Companies' Terms of Service discourage community-led evaluations.}
Many of the findings from the model vulnerability research mentioned in \cref{sec:independent-eval}, such as jailbreaks, bypassing safety guardrails, or text-to-image exploits, are legally prohibited by the terms of service for popular systems, including those of OpenAI, Google, Anthropic, Inflection, Meta, Midjourney, and others. 
While these terms are intended as a deterrent against malicious actors, they also inadvertently restrict safety and trustworthiness research---both by forbidding the research, and enforcing it with account suspensions.
While platforms enforce these restrictions to varying degrees, the terms disincentivize good faith research by granting developers the right to terminate researchers' accounts (without appeal or justification) or even take legal action against them. 
The risk of losing account access may dissuade many researchers altogether, as these accounts are critical for a range of vulnerability and other AI research.

\input{tables/policies}

AI developers’ documentation often purports to support independent research; however, it does not clearly state the conditions under which evaluation and red teaming would not violate the usage policy, leaving researchers uncertain as to whether or how they should conduct their research.
In \cref{tab:interviews}, we share common themes attributed to discussions between ourselves and colleagues, summarizing their experiences conducting evaluation and red teaming research on generative AI platforms.
These themes reflect an imperfect sample: they are skewed in that they represent the opinions of researchers \emph{who chose to conduct safety research}, excluding those who chose not to, lacked access to the companies they would have evaluated, or were deterred for uncertainty of legal liability.

\textbf{Independent AI evaluation is largely inconsistent, opaque, and challenging across companies.} 
From our experience and discussions, the bulk of this research is concentrated on Meta models like Llama-2 \citep{touvron2023llama}, or OpenAI models like ChatGPT \citep{openai2023chatgptapi}.
Llama models are popular as they have downloadable weights, allowing a researcher to red team locally without having their account terminated for usage policy violations. 
OpenAI models are popular as they are accessible via API, are highly performant, and have widespread public use.
While many researchers are tentative about red teaming OpenAI, usage policy enforcement is often lax. 
However, account suspensions in the course of public interest research have taken place, to our knowledge, for each of OpenAI, Anthropic, Inflection, and Midjourney, with Midjourney being the most prolific.
We withhold details on most of these to respect the anonymity of researchers.
As one example, independent evaluation by an artist found Midjourney has a ``visual plagiarism problem'' \citep{marcus2024generative}.
This resulted in their account being repeatedly suspended without warnings or justification.
The cost of suspensions without refunds quickly tallies to hundreds of dollars, and creating new accounts is also not trivial, with blanket bans on credit cards and email addresses.

AI companies have begun using their terms of service to deter analysis, particularly into copyright claims.
Midjourney updated its Terms of Service to include penalties such as account suspension or legal action for conducting such research.\footnote{See \url{https://twitter.com/Rahll/status/1739155446726791470}}
Midjourney's Terms of Service states: ``If You knowingly infringe someone else’s intellectual property, and that costs us money, we’re going to come find You and collect that money from You. We might also do other stuff, like try to get a court to make You pay our legal fees. Don’t do it'' \citep{midjourney2023tos}.\footnote{See Section 10 \url{https://docs.midjourney.com/docs/terms-of-service}}
Llama 2's license will also terminate access if model outputs are used as part of intellectual property litigation.\footnote{See Section 5c: \url{https://ai.meta.com/llama/license/}}

Our analysis of company policies in \cref{tab:policies} shows not all companies disclose their enforcement process (the mechanisms for identifying and enforcing violations of the usage policy).
Google and Inflection are the only companies to provide the user any form of justification on how the usage policy is enforced.
And, only for OpenAI, Inflection, and Midjourney did we find evidence of an enforcement appeals process. 
Without additional information on how companies enforce their policies, researchers have no insight into enforcement appeals criteria, or whether companies reinstate public interest research post-hoc. 

\textbf{Existing safe harbors protect security research but not other good faith research.}
AI developers have engaged to differing degrees with external red teamers and evaluators.
OpenAI, Google, Anthropic, and Meta, for example, have bug bounties, and even safe harbors.
However, companies like Meta and Anthropic currently ``reserve final and sole discretion for whether you are acting in good faith and in accordance with this Policy''. 
They may revoke access rights to models, even open models like Llama 2 \citep{touvron2023llama}, or hold the researchers legally accountable, at their discretion.
This leaves clear ways to stifle and deter good faith research.
Additionally, these safe harbors are tightly-scoped to traditional security issues like unauthorized account access.\footnote{OpenAI expanded its safe harbor to include ``model vulnerability research'' and ``academic model safety research'' in response to an early draft of our proposal, though some ambiguity remains as to the scope of protected activities.}
Developers disallow other model flaws named in their usage policies, including, ``adversarial testing'' \citep{anthropic2023rdp}, ``jailbreaks'', bypassing safety guardrails, or generating hate speech, misinformation, or abusive imagery.

Among other safety research commitments, some companies publish reports on internal evaluation efforts, while others selectively invite third parties to participate in pre-release red teaming, or have researcher access programs for deeper access to released models. 
These are laudable initiatives, especially when they are accompanied by subsidized credits for researchers \citep{openai2024access}. 
Nonetheless, these measures leave significant gaps in the ecosystem for independent evaluations. 
Reports on internal red teaming are often largely irreproducible and generate limited trust due to mismatched corporate incentives (e.g. \citet{anthropic2023frontier}). Invitations to third-party researchers are limited and can be self-selecting. 
And researcher access programs, if available, often do not notify researchers of rejections and thus create an environment of uncertainty \citep{bommasani2023transparency}. 
Researchers have argued that a patchwork of policies like these can create a veneer of open and responsible research, without lifting other obstacles for participatory research \citep{krawiec2003cosmetic,zalnieriute2021transparency,whittaker2021steep}.

Companies should take steps to facilitate independent AI evaluation and reduce the fear of reprisals for safety research. 
The gaps in the policy architectures of leading AI companies, depicted in \cref{tab:policies} force well-intentioned researchers to either wait for approval from unresponsive access programs, or risk violating company policy and potentially losing access to their accounts. 
The net result is a situation akin to companies gatekeeping access to their platforms and thereby restricting the scope of safety research, whether intentional or not. 
This research environment can limit the diversity and representation in evaluation, ultimately stymieing public awareness of risks to AI safety. 

%% file: tables/policies.tex
\begin{table*}[t!]
\centering
\begin{adjustbox}{width=\textwidth}
\begin{tabular}{p{2cm}|p{2.1cm}|p{0.9cm}p{0.9cm}p{0.9cm}p{0.9cm}p{0.9cm}p{0.9cm}p{0.9cm}p{1.3cm}}
\toprule
\textbf{AI Company} & \textbf{AI System} & \rot{Public API / Open} & \rot{Deep Access} & \rot{Researcher Access} & \rot{Bug Bounty} & \rot{Safe Harbor} & \rot{Enforcement Process} & \rot{Enforcement Justification} & \rot{Enforcement Appeal} \\
\midrule
OpenAI & GPT-4 & \bcircle & \wcircle & \bcircle & \bcircle & \wcircle\textsuperscript{\dag} & \bcircle & \ecircle & \wcircle \\
\rowcolor[gray]{0.9} 
Google & Gemini & \bcircle & \ecircle & \ecircle & \bcircle & \ecircle & \ecircle & \wcircle & \ecircle \\
Anthropic & Claude 2 & \ecircle & \ecircle & \wcircle & \ecircle & \wcircle\textsuperscript{\ddag} & \bcircle & \ecircle & \ecircle \\
\rowcolor[gray]{0.9} 
Inflection & Inflection-1 & \ecircle & \ecircle & \ecircle & \ecircle & \ecircle & \ecircle & \wcircle & \wcircle \\
Meta & Llama 2 & \bcircle & \bcircle & \bcircle & \bcircle & \wcircle\textsuperscript{\ddag} & \ecircle & \ecircle & \ecircle \\
\rowcolor[gray]{0.9} 
Midjourney & Midjourney v6 & \ecircle & \ecircle & \ecircle & \ecircle & \ecircle & \ecircle & \ecircle & \wcircle \\
Cohere & Command & \bcircle & \ecircle & \bcircle & \ecircle & \wcircle & \ecircle & \ecircle & \ecircle \\
\bottomrule
\end{tabular}
\end{adjustbox}
\caption{\textbf{A summary of the policies, access, and enforcement for major AI systems, suggesting a challenging environment for independent AI research.} 
We catalog if each system has a public API, deeper access than final outputs (e.g. top-5 logits for OpenAI), researcher access programs, security research bug bounties, any legal safe harbors, and whether they disclose their account enforcement process, disclose justification on enforcement actions, and have an enforcement appeals process.
\bcircle{} indicates the company satisfies this criteria; \ecircle{} indicates it does not, and \wcircle{} indicates partial satisfaction.
\textsuperscript{\ddag} Indicates security-only research safe harbors, ``solely at [their] discretion''.
\textsuperscript{\dag} Indicates a safe harbor for security and ``academic research related to model safety''. The latter was added by OpenAI in response to reading an early draft of this proposal, though some ambiguity remains as to the scope of protected activities.
Full details are provided in \cref{tab:href_policies}.
}
\label{tab:policies}
\vspace{-2mm}
\end{table*}

%% file: icml_sections/04_safe_harbors.tex
\section{Safe Harbors}
\label{sec:safe_harbors}

\begin{figure*}[ht]
    \centering
    \begin{subfigure}{0.94\textwidth}
        \includegraphics[width=\textwidth]{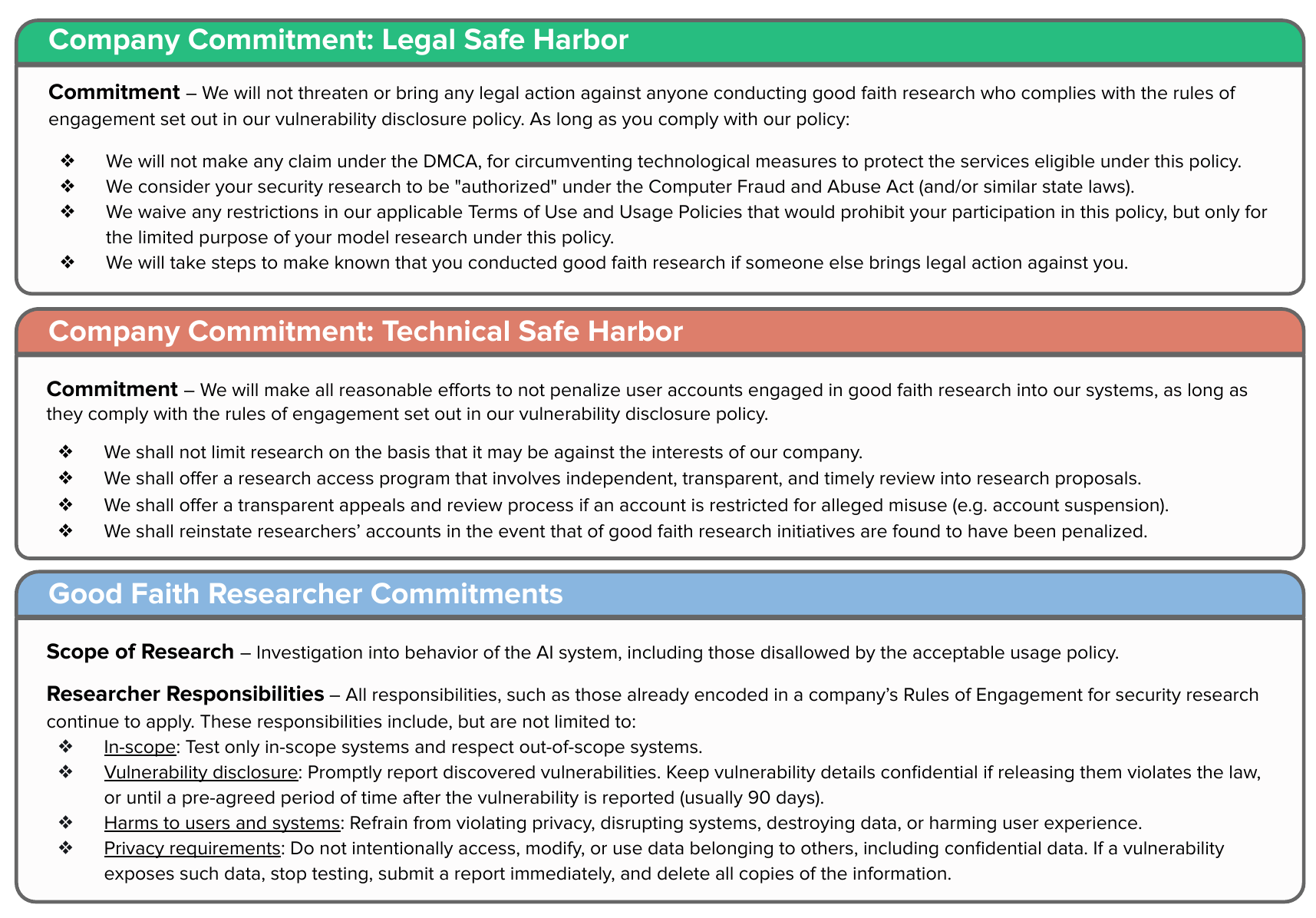}
    \end{subfigure}
    \caption{
    \textbf{A summary of the suggested mutual commitments and scope of a legal safe harbor, and technical safe harbor.} 
    These commitments extend existing safe harbors for security research as well as researcher access programs, and are written in the context of US laws.
    For a wider list of common researcher responsibilities consider \href{https://bugcrowd.com/openai}{OpenAI's Rules of Engagement}.}
    \label{fig:safe-harbor}
    \vspace{-3mm}
\end{figure*}

We believe that a pair of voluntary commitments could significantly improve participation, access, and incentives for public interest research into AI safety. 
The two commitments are: (i) a \textbf{legal safe harbor}, protecting good faith, public interest evaluation research provided it is conducted in accordance with well established security vulnerability disclosure practices, and (ii) a \textbf{technical safe harbor}, protecting this evaluation research from account termination; summarized in \cref{fig:safe-harbor}. 
Both safe harbors should be scoped to include research activities that uncover \textit{any system flaws}, including all undesirable generations currently prohibited by the usage policy. 
As we shall argue later, this would not inhibit existing enforcement against malicious misuse, as protections are entirely contingent on abiding by the law and strict vulnerability disclosure policies, determined ex post.
Existing safe harbor resources \citep{etcovich2018coming, pfefferkorn2022shooting, hackerone2023safeharbor}, and vulnerability disclosure policies \citep{google2010rebooting,bugcrowd2023vulnerability} provide grounding for these proposals.
In particular, \citet{elazari2018hacking,elazari2019private,akgul2023bug,kenway2022bugs} discuss the implementations of algorithmic bug bounties, \citet{walshe2023towards} note ambiguities on formal constraints, and \citet{raji2022audit} explore governance for third-party AI audits, including legal protections for researchers.
The legal safe harbor, similar to the proposal by \citet{abdo2022safe} for social media platforms, would safeguard certain research from some amount of legal liability, mitigating the deterrent of strict terms of service and the threat that researchers’ actions could spark legal action by companies (e.g. under US laws such as the CFAA or DMCA Section 1201). 
The most important condition of a legal safe harbor is the determination of acting in good faith should not be ``at the sole discretion'' of the companies, as Meta and Anthropic have currently defined it.
The technical safe harbor would limit the practical barriers erected by usage policy enforcement, with consistent and broader community access for important, public interest research.
Together these steps would reduce the legal and practical obstacles to conducting independent evaluation and red teaming research.

\subsection{A Legal Safe Harbor}

A legal safe harbor could mitigate risks from civil litigation, providing assurances that AI platforms will not sue researchers if their actions were taken for research purposes. 
Take, for example, the U.S. legal regime, which governs many of the world's leading AI developers.
The Computer Fraud and Abuse Act (CFAA), which allows for civil lawsuits for accessing a computer without authorization or exceeding authorized access \citep{cfaa}, could be used by AI developers to sue researchers for accessing their models in a way that was unintended, though there are complexities to the legal analysis for adversarial attacks on AI models~\citep{calo2018tricking}. 
Section 1201 of the Digital Millennium Copyright Act (DMCA) allows for civil lawsuits if researchers circumvent technological protection measures (TPMs), which effectively control access to works protected by copyright \citep{dmca}. 
These risks are not theoretical; security researchers have been targeted under the CFAA \citep{pfefferkorn2021cfaa}, and DMCA § 1201 hampered security researchers to the extent that they requested a DMCA exemption for this purpose \citep{colannino2021dmca}.
Already, in the context of generative AI, OpenAI has attempted to dismiss the New York Times v OpenAI lawsuit \citep{GrynbaumMac2023} on the allegation that New York Times research into the model constituted hacking \citep{Brittain2024}.
Relatedly, a petition for an exemption to the DMCA has been filed requesting that researchers be allowed to investigate bias in generative AI systems \citep{weiss2023petition}.

\citet{abdo2022safe} argue a safe harbor is oriented around conditions of access, rather than \emph{who} gets access.
The protections apply only to parties who abide by the rules of engagement, to the extent they can subsequently justify their actions in court.
Typically, responsible vulnerability disclosure policies impose strict criteria for when the vulnerability should be disclosed, how long before it can be released to the public, privacy protection rules, and other criteria for the most dangerous exploits.
Research that strays from those reasonable measures, or is already illegal, would not succeed in claiming those protections in an ex post investigation. 
As such, malicious use would remain legally deterred, and platforms would still be obligated to prevent misuse. 
\citet{abdo2022safe} argue a safe harbor designed in this way, based on ex post researcher conduct, would not enable malicious use any more than in its absence. 
Nor would it alter platforms' obligations to protect their users against third parties or from enforcing malpractice.

Companies’ legal safe harbors would protect researchers from civil liability, not criminal liability.
Knowingly querying a model to generate certain types of content, whether for red teaming or not, can be illegal in certain jurisdictions---particularly in the case of image- or video-generation systems \citep{gupta2024laion}. 
Moreover, certain violations of DMCA § 1201, particularly those that are committed ``willfully and for purposes of commercial advantage or private financial gain,'' can lead to criminal liability \citep{dmca1204a}, as can many violations of the CFAA. 
We would recommend governments provide clear guidelines and, where appropriate, safe harbors for safe and responsible red teaming of illegal content generated by models. 
Such safe harbors against criminal conduct may need to be codified into statute in order to be guaranteed.
However, they could be implemented by statements of policy, for example, such as when the Department of Justice issued a new policy in 2022 stating that ``good-faith security research should not be charged'' \citep{doj2022cfaa}.

The US Executive Order on AI directs the National Institute of Standards and Technology (NIST) to establish guidelines for conducting red-teaming and assessing the safety of foundation models \citep{EO14110}. Standardizing a legal safe harbor for researchers would complement NIST’s comprehensive AI evaluation agenda and its AI Risk Management Framework \citep{nist2024EOeval, nist2023airmf}. The US AI Safety Institute Consortium, a public-private research collaboration, could be used to promote the adoption of safe harbors among companies \citep{nist2023consortium}.

\subsection{A Technical Safe Harbor}
\label{sec:technical-safe-harbor}

Legal safe harbors still do not prevent account suspensions or other enforcement action that would impede independent safety and trustworthiness evaluations. 
Without sufficient technical protections for public interest research, a mismatch can develop between malicious and non-malicious actors since the latter are discouraged from investigating vulnerabilities exploited by the former.
We propose companies offer some path to eliminate these technical barriers for good faith research. 
This would include more equitable opportunities for researcher access, and guarantees that those opportunities will not be foreclosed for researchers who adhere to companies' guidelines.

The challenge with implementing a technical safe harbor is distinguishing between legitimate research and malicious actors, without notable costs to developers. 
An exemption to usage moderation may need to be reviewed in advance, or at least when an unfair account suspension occurs.
However, we believe this problem is tractable, and offer recommendations, grounded in prior proposals.
First, we discuss how to scale up participation by delegating responsibilities to trusted independent third parties to \emph{pre-review} researcher access.
Then we discuss how an independently reviewed and transparent account suspension appeals process could enable fairer \emph{post-review} to researcher access.
Independent review and scaling participation are staples of both options.

\textbf{Independent third parties like universities or NAIRR can scale participation in AI evaluation, without misaligned corporate incentives.}
To facilitate more equitable access, and reduce the potential for corporate favoritism, we propose the responsibility of access authorization be delegated to trusted third parties, such as universities, government, or civil society organizations. 
The U.S. National Artificial Intelligence Research Resource (NAIRR) offers a suitable vehicle for a pilot of this approach as it already partners with and shares resources between AI developers and nonprofits. 
AI developers provide resource credits through NAIRR, and OpenAI has called for wider participation: ``by providing broader access to essential tools and data, we are opening doors for a diverse range of talents and ideas, furthering innovation and ensuring that AI development continues to be a force for the greater good'' \citep{nsf2024ai}.

A similar approach has already been adopted to provide independent access to Meta’s social media user data, with the University of Michigan as the trusted intermediary \citep{gonzalez2023fb}. 
This solution scales, with partner organizations likely to aid in access review in exchange for wider participation in AI red teaming. 
It also effectively diverts responsibility from corporate interests to organizations already invested in fair, responsible, and accountable AI research. 
These partnerships do not require AI developers to fully relinquish access control but are a meaningful step in facilitating more equitable access without stretching their own resources.
Each partner organization’s API usage could be traced to their API keys---essentially a ``researcher API''. 
Organizations would have autonomy to authorize their own network of researchers, but would be responsible for any misuse tied to their API keys. 

A number of similar proposals, discussed in \cref{sec:related-proposals}, have been made for independent researcher access, like structured access or review boards, both of which would delegate the responsibility of access selection to independent third parties.
While this approach scales well and adopts independent access privileges, it can have severe limitations if AI companies only select a very finite set of partners, or choose to exclude more critical organizations.
As a start, we recommend allowing NAIRR to help formulate the partner network, to include a set of trusted international academic organizations, as well as nonprofits in NAIRR such as AI2, EleutherAI, and MLCommons.
Already these changes would make significant strides in expanding access through independent review.

\textbf{Transparent access and appeals processes can improve community trust.}
Some generative AI companies may be unwilling to share access authorization more widely. 
There is a clear alternative: commit to a transparent access appeals process that makes decision criteria and outcomes visible to the wider community.
Ideally, this process would be reviewed independently, perhaps with the help of NAIRR partner organizations.
Whenever public interest evaluation research is suspended, researchers should have the opportunity to appeal the decision under a technical safe harbor. 
Companies can adopt an access process with clearly codified selection criteria, guaranteeing they will respond to applicants within a certain period of time, with a justification for the outcome decision. 
While this would not address the need for additional resources, it would provide the AI community with significantly greater visibility into companies' decisions to grant access, and allow the community to apply collective pressure against any attempt to restrict legitimate research.
The common denominator between pre-review and post-review technical safe harbors, described above, is providing a fair process to enable good faith research without the fear of unjustified account suspensions.
In \cref{sec:technical-safe-harbor-approach} we sketch an implementation of a pre-registration and appeals process, based on existing researcher access programs, that could facilitate implementation of a technical safe harbor.

There are many dimensions of improving researcher access, including earlier access, deeper access, and subsidized access. 
The technical safe harbor described is a precondition for more independent and broader participation across all these axes, should companies offer earlier, deeper, or subsizided access.
While efforts by AI companies to broaden safety research, such as accepting community applications for pre-release red-teaming and subsidizing such research with compute credits are useful first steps, the safe harbors we propose would strengthen broader research protections while being more independent of AI companies' control.

%% file: icml_sections/05_related_proposals.tex
\section{Related Proposals}
\label{sec:related-proposals}

Our proposals for legal and technical safe harbors build on prior calls to expand independent access for AI evaluation, red teaming, and safety research. 
\citet{hpc2023legal} has proposed that governments ``clarify and extend legal protections for independent AI red teaming,'' similar to our voluntary legal safe harbor proposal. 
The Council stated, ``the same industry norms on providing time to mitigate before public disclosure, and avoiding retaliation for good faith disclosures, should eventually apply to AI misalignment disclosures as they do for security vulnerability disclosures.'' 
The Algorithmic Justice League has advocated for vulnerability disclosure for algorithmic harms, calling for independent algorithmic audits involving impacted communities \citep{costanza2022auditauditors, kenway2022bugs}.
Moreover, AI Village hosts events where large groups of independent researchers red team generative AI models for a wide range of vulnerabilities \citep{cattell2023recap}.
An array of researchers have recommended additional external scrutiny of the emerging risks and overall safety of frontier AI models to ``improve assessment rigor and foster accountability to the public interest'' \citep{anderljung2023frontier}.
\citet{bucknall2023structured} have also proposed structured access for third party research via a dedicated research access API, with third-party independent review. 
Stanford's Center for Research on Foundation Models has proposed an independent Foundation Models Review Board to moderate and review requests for deeper researcher access to foundation models \citep{liang2022norms}.

Governments have also suggested the need for independent evaluation and red teaming. 
The US Office of Management and Budget's Proposed Memorandum on Advancing Governance, Innovation, and Risk Management for Agency Use of Artificial Intelligence encourages federal agencies to consider as part of procurement contracts for generative AI systems ``requiring adequate testing and safeguards, including external AI red teaming, against risks from generative AI such as discriminatory, misleading, inflammatory, unsafe, or deceptive outputs'' \citep{omb2023ai}.
The EU AI Act states that providers of general-purpose AI models with systemic risks must share a ``detailed description of the measures put in place for the purpose of conducting internal and/or external adversarial testing (e.g. red teaming), model adaptations, including alignment and fine-tuning'' to the EU as part of their technical documentation \citep{eu2023aiact, hacker2024final}.
In addition, Canada's Voluntary Code of Conduct on the Responsible Development and Management of Advanced Generative AI Systems includes a commitment that developers will ``conduc[t] third-party audits prior to release'' \citep{canada2023vc}.

%% file: icml_sections/06_conclusion.tex
\section{Conclusion}
\label{sec:conclusion}

The need for independent AI evaluation has garnered significant support from academics, journalists, and civil society.
Examining challenges to external evaluation of generative AI systems, we identify legal and technical safe harbors as minimum and fundamental protections.
We believe they would significantly improve norms in the ecosystem and drive more inclusive community efforts to tackle the risks of generative AI.

%% file: icml_sections/10_appendix.tex
\section{Additional Considerations \& Future Work}
\label{sec:future-work}

There are a number of future research directions that would help in making a safe harbor for AI evaluation and red teaming a reality. 
For instance, our proposal would benefit from further exploration of some of the challenging aspects in designing a technical safe harbor.
In particular, to agree to such a commitment, AI companies will be concerned with protecting their own intellectual property and sensitive data.
While restrictions on publicizing these valuable assets are often included in standard vulnerability disclosure policies, there is an implicit tension between expanding access to a greater number of independent researchers and ensuring compliance with disclosure policies.
As AI models also expose new risks and harms, the definitions of ``good faith'' research may need to be flexible and evolve.

Our safe harbor proposals are formulated within the context of the US legal system.
It is likely that different jurisdictions impose substantially different legal requirements related to research on the safety, security, and trustworthiness of AI.
The use of geo-location in social media and search engines has allowed for digital platforms to tailor the behavior of their algorithmic systems based on each region.
Generative AI companies may also adopt geo-location to customize their policies and enforcement of those policies by region.
Future work should consider these changes and how a safe harbor proposal could work to achieve its aims in supporting fair, transparent, and inclusive good faith research internationally.

This line of research would also benefit from a more robust engagement with counterarguments to these proposals. While we believe the benefits of wider participation in independent AI safety and trustworthiness research will outweigh any risks to misuse, especially for well designed safe harbors, others may disagree.
These trade-offs deserve more empirical analysis to understand the effects of such proposals.

\section{Details on Access \& Enforcement Policies}
\label{sec:details}

\input{tables/circle-policies-href}

In \cref{tab:policies} we summarize the policies, access, and enforcement for the major AI companies and their flagship systems.
In \cref{tab:href_policies} we link the evidence for each determination.
And in this section we describe the criteria for each column in greater detail.

\begin{itemize}[itemsep=0pt]
    \item \textbf{Usage Policy:} \bcircle{} indicates that the company documents its acceptable usage policy, which they all do.
    
    \item \textbf{Deep Access:} \bcircle{} indicates that the company provides some level of access to the AI system in question (OpenAI provides the top 5 logits and Meta provides open weights), \ecircle{} indicates there is no deeper access to the model (as is the case for all other companies). 
    
    \item \textbf{Researcher Access:} \bcircle{} indicates that the company maintains a researcher access program (OpenAI, or Meta with released model weights), \wcircle{} indicates there is some access for researchers with some caveats (Anthropic has a limited early access program), \ecircle{} indicates there is no researcher access (as is the case for all other companies). 

    \item \textbf{Safe Harbor:} \bcircle{} would indicate that there is a legal safe harbor for model vulnerabilities beyond security research. \wcircle{} indicates there is form of commitment to research exemptions. OpenAI, Anthropic and Meta have a safe harbor only for security research. \ecircle{} indicates there is no safe harbor (all other companies).
    OpenAI's new safe harbor (since updating in late January, in response to this proposal) is the closest to a full legal safe harbor, though there remains some ambiguity remains as to the scope of protected activities.
    For Cohere, while it does not have a safe harbor, their usage policy says ``Note about adversarial attacks: Intentional stress testing of the API and adversarial attacks are allowable, but violative generations must be disclosed here, reported immediately, and must not be used for any purpose except for documenting the result of such attacks in a responsible manner.''
    Meta also provides a similar safe harbor for \emph{in-scope} activities, which appear to be ``integral privacy or security issues associated with Meta's large language model, Llama 2, including being able to leak or extract training data through tactics like model inversion or extraction attacks.''
    However, like Anthropic, it's safe harbor is determined at their sole discretion, and therefore provides limited benefit.
    
    \item \textbf{Enforcement process:} \bcircle{} indicates that the company shares significant detail about how it enforces its usage policy such as the specific practices it uses for enforcement (OpenAI, Anthropic), \ecircle{} indicates there is little or no detail publicly available about the specific ways that the company enforces its usage policy (all other companies). 
    Each company prescribes a prohibited set of uses, required by their terms of service, and all of these are enforced with moderation systems in the APIs and playgrounds, though only OpenAI and Anthropic openly disclose this. For instance, in GPT-4’s System Card OpenAI acknowledges using “a mix of reviewers and automated systems to identify and enforce against misuse”, and that policy-violating content will trigger warnings, suspensions and bans.

    \item \textbf{Enforcement justification:} \bcircle{} would indicate that the company provides a specific reason for why a certain prompt or query was violative, \wcircle{} indicates that the company provides some detailed (if non-specific) justification when a user's prompt or query is blocked or otherwise deemed violative (Google, Inflection), \ecircle{} indicates there is no significant justification provided (all other companies). 

    \item \textbf{Enforcement Appeal:} \bcircle{} indicates that the company provides an appeals process when it takes an enforcement action under its usage policy (OpenAI, Inflection, Midjourney), \ecircle{} indicates there is no appeals process (all other companies).
\end{itemize}

\section{Implementation of a Technical Safe Harbor}
\label{sec:technical-safe-harbor-approach}

In \cref{sec:technical-safe-harbor} we discuss two approaches by which companies can establish a technical safe harbor---by scaling researcher participation and enlisting independent judgement of what constitutes good faith research, without taxing corporate resources. 
These approaches offer two lenses: pre-review of research applications or post-review of suspended researchers.
In reality, some combination of the two may be most convenient and efficient.
Here we sketch a proposal for an independently reviewed appeals process (post-review), but that requires research pre-registration to ease the challenge of reviewing whether research is good faith.
A key choice is to determine the set of acceptable institutions for research pre-registration, which would ideally be negotiated ahead of time with NAIRR.
We sketch what the components of this system might look like:

\begin{itemize}[itemsep=0pt]
    \item \textbf{Good Faith Research Pre-Registration:} Good faith researchers can pre-register their work, establishing in advance their affiliations, intent, and research goals, so the company can easily cross-reference flagged accounts with these detailed forms. Similar to the existing \href{https://openai.com/form/researcher-access-program}{OpenAI Researcher Access Program}, or Twitter's 2021 Researcher API (before it was decommisioned), the pre-registration form can include: Name, API key, institutional affiliations, evidence of affiliation (email and website), list of investigators, intended research focus, specific sensitive topics that violate the usage policy, timeline, etc.
    \item \textbf{Vulnerability Disclosure:} The researchers should tag vulnerability disclosures through the same platform, so these can be directly connected to the pre-registration form.
    \item \textbf{Criteria for Technical Safe Harbor:} If an account is flagged to a company, either because it violated its usage policy, or for some other reason, the company can directly cross-reference the account with pre-registered forms. If a pre-registration does not exist, the company can suspend the account. If a pre-registration form does exist, the company can review the account's eligibility for an exemption from enforcement based on a number of factors: (i) is the account affiliated with a recognized academic or research institution, (ii) are the usage policy violations in line with the proposed research topics/timeline, and (iii) is there any evidence that the researcher has violated the vulnerability disclosure policy, such as publishing vulnerabilities without advance disclosure (in the required timeframe).
    We recommend that acceptable research institutions be negotiated in advance under the guidance of NAIRR.
    Ideally the group of acceptable research institutions would include major international universities as well as organizations with a track record for trusted research, such as AI2, EleutherAI, and Masakhane.
    In the event that each of these criteria are met and the company still has concerns, it can suspend the account and then directly contact the organization or supervisor of the work, as disclosed in the form, with the justification for suspension.
    \item \textbf{Suspension Appeals Process:} If the account is suspended, despite the researcher having pre-registered their research plan, there may be an incongruity or ambiguity in their application.
    The account holder will have the option to appeal this process, ideally with an impartial, independent reviewer.
    If necessary, the company could escalate the appeal to the university or organization's department leads, to ensure the organization stands by the researcher's work.
    This would likely rule out the vast majority of malicious actors, and distribute the responsibility between AI companies and research institutions themselves.
    The appeals process should have standardized, well-documented criteria and a fair timeline (e.g. 30 days). 
\end{itemize}

\section{Company Support for Wider Participation in AI Evaluations}
\label{sec:company-support}

There is ample evidence that prominent AI companies are verbally committed to independent and broader AI system evaluations.
OpenAI's Sharing \& Publication Policy states ``we believe it is important for the broader world to be able to evaluate our research and products, especially to understand and improve potential weaknesses and safety or bias problems in our models'' \citep{openai2023sharing}.
It remains unclear how this commitment relates to OpenAI's terms of service and their enforcement.\footnote{We emailed ``papers@openai.com'' to ask for clarification on research exemptions for the OpenAI Usage Policy, but received no response.}
Anthropic has stated in its  Core Views on AI Safety that ``in the near future, we also plan to make externally legible commitments to only develop models beyond a certain capability threshold if safety standards can be met, and to allow an independent, external organization to evaluate both our model’s capabilities and safety'' \citep{anthropic2023core}.
As part of its Secure AI Framework, Google has committed to ``Expanding our bug hunters programs (including our Vulnerability Rewards Program) to reward and incentivize research around AI safety and security'' \citep{google2023SAIF}.
Meta has highlighted the importance of external red teams in improving the safety of Llama 2, noting that ``Our extensive testing through both internal and external red teaming is continuing to help improve our AI work across Meta'' \citep{meta2023safetypolicies}.
In the same vein, Inflection states ``Red-teaming is and will continue to be the engine at the heart of our evaluation framework. Red-teams provide the best indication of how a model will perform in real-world situations ... To do this, we commission outside experts as well as relying on our safety team. Inflection is currently building teams of highly specialized red-teamers that can bring their unique expertise to investigate models in a manner our `in-house' teams would not have the context to do effectively'' \citep{inflection2023safetypolicies}.
The \href{https://www.frontiermodelforum.org/}{Frontier Model Forum}, comprised of OpenAI, Google, Anthropic, and Microsoft, states that one of its core objectives is ``Advancing AI safety research ... Research will help promote the responsible development of frontier models, minimize risks, and enable independent, standardized evaluations of capabilities and safety.''

\section{Additional Red Teaming Work}
\label{app:more-red-teaming}

In addition to the works on AI audits, red teaming, and evaluations cited in \cref{sec:eval}, there are many other notable works, worthy of further discussion.
\citet{shah2023scalable} find GPT-4 will give instructions for making weapons and narcotics.
\citet{fang2024llm} shows how GPT-4 can be used to automatically hack websites in the right circumstances.
\citet{sharma2023towards} discuss the behavior of model sycophancy.
\citet{santurkar2023whose} show political and ideological biases systemic in AI models.
\citet{ji2023survey,zhang2023language} demonstrate the challenges with model hallucination.
\citet{qu2023unsafe} illustrates models' capacities for harmful content generation.
Lastly, \citet{rando2022red} red teams Stable Diffusion's safety filters, revealing flaws.








%% file: tables/circle-policies-href.tex
\begin{table*}[t!]
\centering
\begin{adjustbox}{width=\textwidth}
\begin{tabular}{p{2cm}|p{2.1cm}|p{0.9cm}p{0.9cm}p{0.9cm}p{0.9cm}p{0.9cm}p{0.9cm}p{0.9cm}p{1.3cm}}
\toprule
\textbf{AI Company} & \textbf{AI System} & \rot{Usage Policy} & \rot{Deep Access} & \rot{Researcher Access} & \rot{Bug Bounty} & \rot{Safe Harbor} & \rot{Enforcement Process} & \rot{Enforcement Justification} & \rot{Enforcement Appeal} \\
\midrule
OpenAI & GPT-4 & \bcircle \href{https://openai.com/policies/usage-policies}{[Link]} & \wcircle \href{https://platform.openai.com/docs/introduction}{[Link]} & \bcircle \href{https://openai.com/form/researcher-access-program}{[Link]} & \bcircle \href{https://bugcrowd.com/openai}{[Link]} & \wcircle \href{https://bugcrowd.com/openai}{[Link]} & \bcircle  \href{https://arxiv.org/abs/2303.08774}{[Link]} & \ecircle \href{https://github.com/stanford-crfm/fmti/blob/main/scoring/OpenAI%202023%20FMTI%20Scores.pdf}{[Link]} & \wcircle \href{https://github.com/stanford-crfm/fmti/blob/main/scoring/OpenAI%202023%20FMTI%20Scores.pdf}{[Link]} \\
\rowcolor[gray]{0.9} 
Google & Gemini & \bcircle \href{https://web.archive.org/web/20230914001155/https://policies.google.com/terms/generative-ai/use-policy}{[Link]} & \ecircle & \ecircle & \bcircle \href{https://security.googleblog.com/2023/10/googles-reward-criteria-for-reporting.html}{[Link]} & \ecircle \href{https://bughunters.google.com/about/rules/6625378258649088/google-and-alphabet-vulnerability-reward-program-vrp-rules}{[Link]} & \ecircle \href{https://github.com/stanford-crfm/fmti/blob/main/scoring/Google%202023%20FMTI%20Scores.pdf}{[Link]} & \wcircle \href{https://ai.google.dev/docs/safety_setting_gemini}{[Link]} & \ecircle \href{https://github.com/stanford-crfm/fmti/blob/main/scoring/OpenAI%202023%20FMTI%20Scores.pdf}{[Link]} \\
Anthropic & Claude 2 & \bcircle \href{https://console.anthropic.com/legal/aup}{[Link]} & \ecircle & \wcircle \href{https://www.anthropic.com/earlyaccess}{[Link]} & \ecircle & \wcircle \href{https://www.anthropic.com/responsible-disclosure-policy}{[Link]} & \bcircle \href{https://console.anthropic.com/legal/aup}{[Link]} & \ecircle \href{https://github.com/stanford-crfm/fmti/blob/main/scoring/Anthropic%202023%20FMTI%20Scores.pdf}{[Link]} & \ecircle \href{https://github.com/stanford-crfm/fmti/blob/main/scoring/Anthropic%202023%20FMTI%20Scores.pdf}{[Link]} \\
\rowcolor[gray]{0.9} 
Inflection & Pi & \bcircle \href{https://pi.ai/profile/terms}{[Link]} & \ecircle & \ecircle & \ecircle & \ecircle & \ecircle \href{https://github.com/stanford-crfm/fmti/blob/main/scoring/Inflection%202023%20FMTI%20Scores.pdf}{[Link]} & \wcircle \href{https://heypisupport.zendesk.com/hc/en-us/articles/17791183959437-Understanding-Account-Suspension-Why-was-my-account-suspended-}{[Link]} & \wcircle \href{https://web.archive.org/web/20230914081031/https://docs.google.com/forms/d/e/1FAIpQLScv5_-e6HjkvnqRBrpx8OG65PSAkX9XLXVMnE5eTpOdZQxF3Q/viewform}{[Link]} \\
Meta & Llama 2 & \bcircle \href{https://ai.meta.com/llama/use-policy/}{[Link]} & \bcircle\href{https://llama.meta.com/}{[Link]} & \bcircle\href{https://llama.meta.com/}{[Link]} & \bcircle \href{https://www.facebook.com/whitehat/info/}{[Link]} & \wcircle \href{https://www.facebook.com/whitehat/info/}{[Link]} & \ecircle \href{https://github.com/stanford-crfm/fmti/blob/main/scoring/Meta%202023%20FMTI%20Scores.pdf}{[Link]} & \ecircle \href{https://github.com/stanford-crfm/fmti/blob/main/scoring/Meta%202023%20FMTI%20Scores.pdf}{[Link]} & \ecircle \href{https://github.com/stanford-crfm/fmti/blob/main/scoring/Meta%202023%20FMTI%20Scores.pdf}{[Link]} \\
\rowcolor[gray]{0.9} 
Midjourney & Midjourney v6 & \bcircle \href{https://docs.midjourney.com/docs/terms-of-service}{[Link]} & \ecircle & \ecircle & \ecircle & \ecircle & \ecircle & \ecircle & \wcircle \href{https://docs.midjourney.com/docs/terms-of-service}{[Link]} \\
Cohere & Command & \bcircle \href{https://docs.cohere.com/docs/usage-guidelines}{[Link]} & \ecircle & \bcircle \href{https://txt.cohere.com/c4ai-research-grants/}{[Link]}  & \ecircle & \wcircle \href{https://docs.cohere.com/docs/usage-guidelines}{[Link]} & \ecircle \href{https://github.com/stanford-crfm/fmti/blob/main/scoring/Cohere%202023%20FMTI%20Scores.pdf}{[Link]} & \ecircle \href{https://github.com/stanford-crfm/fmti/blob/main/scoring/Cohere%202023%20FMTI%20Scores.pdf}{[Link]} & \ecircle \href{https://github.com/stanford-crfm/fmti/blob/main/scoring/Cohere%202023%20FMTI%20Scores.pdf}{[Link]} \\ 
\bottomrule
\end{tabular}
\end{adjustbox}
\caption{\textbf{A summary of the policies, access, and enforcement for major AI systems, with links to evidence where applicable.} 
\bcircle{} indicates that a company satisfies or provides access to information in a column, \ecircle{} indicates it does not, and \wcircle{} indicates partial satisfaction. 
}
\label{tab:href_policies}
\end{table*}